\documentclass{article}
\usepackage{spconf,amsmath,graphicx}
\usepackage{booktabs}
\usepackage{diagbox} 

\usepackage{breqn}
\usepackage{graphicx}
\usepackage{subcaption}
\usepackage{booktabs}
\usepackage{diagbox} 
\usepackage{amssymb}
\usepackage{pifont}
\usepackage{textcomp}
\usepackage{xcolor}
\usepackage{nicefrac}

\usepackage{multirow}
\usepackage{makecell}
\usepackage[numbers,sort]{natbib}

\def\newpara{\vspace{1pt}}
\def\psec{\vspace{-4pt}}

\newcommand{\cmark}{\ding{51}}%
\newcommand{\xmark}{\ding{55}}%

\title{ASR is all you need: cross-modal distillation for lip reading}
\name{Triantafyllos Afouras$^1$, Joon Son Chung$^{1,2}$, Andrew Zisserman$^1$}
\address{$^1$ Visual Geometry Group, Department of Engineering Science, University of Oxford \\
$^2$ Naver Corporation}
\begin{document}
%
\maketitle
\begin{abstract}
The goal of this work is to train strong models for visual speech recognition
without requiring human annotated ground truth  data.
We achieve this by distilling from an Automatic Speech Recognition (ASR) model that has been trained on a large-scale audio-only corpus.
We use a cross-modal distillation method that combines Connectionist Temporal Classification (CTC) with a frame-wise cross-entropy loss.
Our contributions are fourfold:
(i) we show that ground truth transcriptions are not necessary to train a lip reading system;
(ii) we show how arbitrary amounts of unlabelled video data can be leveraged to improve performance;
(iii) we demonstrate that distillation significantly speeds up training;
and, (iv) we obtain state-of-the-art results on the
challenging LRS2 and LRS3 datasets for training only on publicly available data.
\end{abstract}
\psec
\psec
\section{Introduction}
\label{sec:intro}
\psec


Visual speech recognition (VSR) has received increasing amounts of
attention in recent years due to the success of deep learning
models trained on corpora of aligned text and face videos~\cite{Assael16,Chung16,Chung17}.
In many machine learning applications, training on very large datasets has proven to have
huge benefits,  and indeed~\cite{Shillingford18,makino2019recurrent} recently demonstrated significant performance
improvements by training on very large-scale proprietary datasets.
However, the largest publicly available datasets for training and evaluating visual speech
recognition, LRS2 and LRS3~\cite{Chung17,Afouras18d},  are orders of magnitude smaller than
their audio-only counterparts used for training Automatic Speech Recognition (ASR) models~\cite{panayotov2015librispeech,Baumann2018}.
This indicates that there are potential gains to be made from a scalable method that
could exploit vast amounts of unlabelled video data.

In this direction, we propose to train a VSR model by {\em distilling} from an ASR model with a
teacher-student approach.
This opens up the opportunity to train VSR model on audio-visual datasets that are an order of magnitude larger
than LRS2 and LRS3, such as VoxCeleb2~\cite{Chung18a} and AVSpeech~\cite{ephrat2018looking}, 
but  lack text annotations. More generally, the VSR model can be trained from {\em any} available video of
talking heads, e.g.\ from YouTube. Training by distillation eliminates the need for
professionally transcribed subtitles, and also removes the costly step of
forced-alignment between the subtitles and  speech required to create VSR training data~\cite{Chung16}.

\begin{figure}[!thb]
\centering 
\vspace{-10pt}                              
\includegraphics[width=1\linewidth]{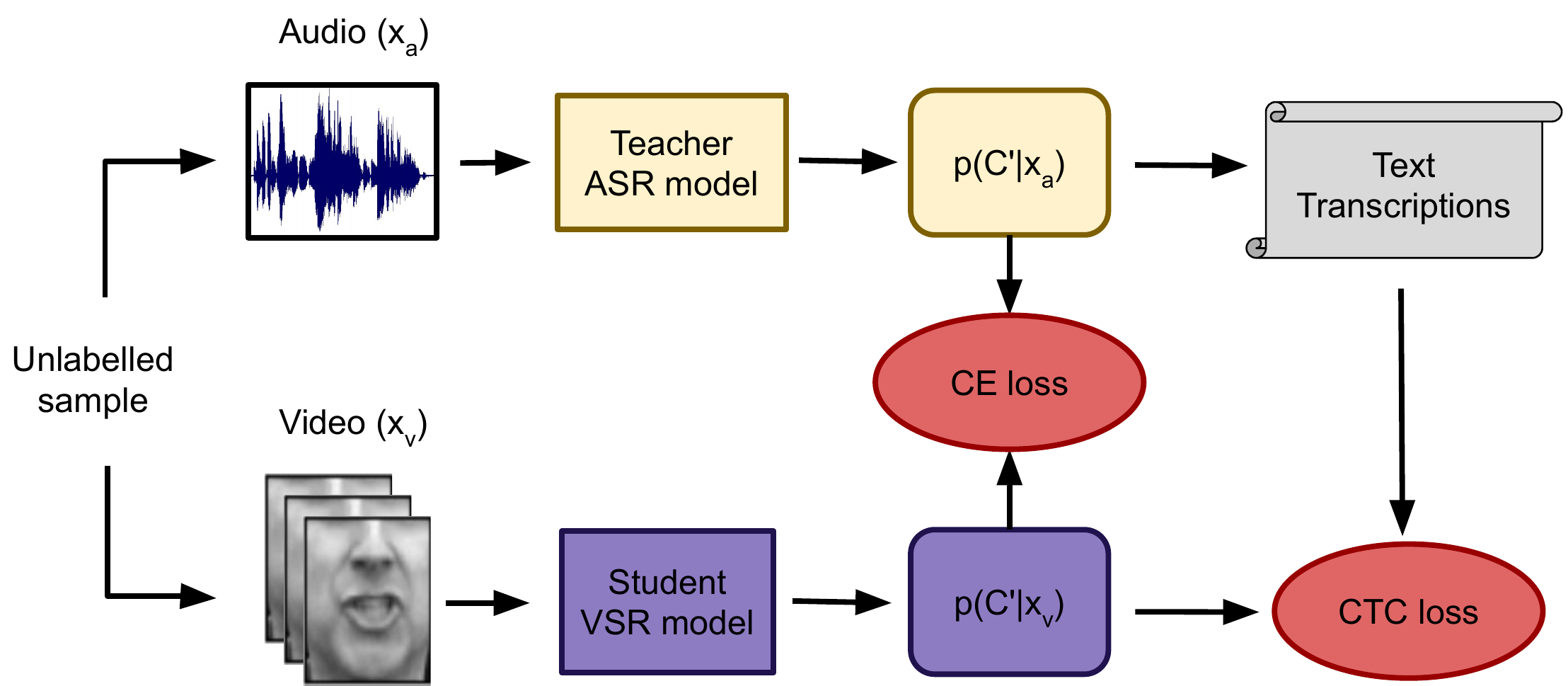}
\caption{Cross-modal distillation of an ASR teacher into a student VSR model.
  CTC loss on the ASR-generated transcripts is combined with minimizing the
  KL-divergence between the student and teacher posterior distributions. 
}
\label{fig:teaser} 
\vspace{-15pt}                              
\end{figure}

Our aim is to to pretrain on large unlabelled datasets in order to boost lip reading performance.
In the process we also discover that human-generated captions are actually not necessary to train a
good model. 
The approach we follow, as shown in Fig.~\ref{fig:teaser}, combines a distillation loss with conventional 
Connectionist Temporal Classification (CTC)~\cite{Graves06}.
An alternative option to exploit the extra data, would have been to train solely with
CTC on the
ASR transcriptions. However we find that compared to that approach, distillation provides a
significant acceleration to training.

\psec
\psec
\subsection{Related Work}
\psec

\newpara\noindent{\textbf{Supervised lip reading.}}
There have been a number of recent works on lip reading using
datasets such as LRS2~\cite{Chung17} and LRS3~\cite{Afouras18d}. 
Works on word-level lip reading~\cite{Chung16} have proposed CNN models and temporal fusion methods
for word-level classification. \cite{Stafylakis17} combines a deeper residual
network and an LSTM classifier to achieve the state-of-the-art on the same task.
Of more relevance to this work is open set character-level lip reading, for which recent work can be divided into two groups. 
The first uses CTC where the model predicts frame-wise labels and is trained
to minimize the loss resulting from all possible input-output alignments under a monotonicity constraint.
LipNet~\cite{Assael16} and more recently~\cite{Shillingford18,makino2019recurrent} are based on this
approach. \cite{makino2019recurrent} in particular demonstrates state-of-the-art performance by training on proprietary data that is orders of magnitude larger than any public dataset. 
The second group is sequence-to-sequence models that 
predict the output sequence one token at a time in an autoregressive manner, attending
to different parts of the input sequence on every step. 
Some examples are the sequence-to-sequence LSTM-with attention model used by~\cite{Chung17} and the
Transformer-based model used by~\cite{Afouras18b} or a convolutional variant by~\cite{zhang2019spatio}.
\cite{petridis2018audio,Afouras19} take a hybrid approach that combines the two ideas, namely using a CTC loss
with attention-based models.
Both approaches can use external language models during inference to boost
performance\cite{Kannan17,Maas15}

\newpara\noindent{\textbf{Knowledge distillation (KD).}}
Distilling knowledge between two neural networks has been popularised
by~\cite{hinton2015distilling}. Supervision provided by the teacher is used to train the student on
potentially unlabelled data, usually from a larger network into a smaller network to reduce model size. 
There are two popular ways of distilling information: training the student to regress the teacher's
pre-softmax logits~\cite{ba2014deep}, and minimising the cross-entropy between the probability outputs~\cite{li2014learning,hinton2015distilling}. 

\newpara\noindent{\textbf{Sequence and CTC distillation.}} 
KD has also been studied in the context of sequence modeling.
For example it has been used to compress sequence-to-sequence
models for neural machine translation~\cite{kim2016sequence} and ASR~\cite{kim2019knowledge}.
Distillation of acoustic models trained with CTC has also been investigated for
distilling a BLSTM model into a uni-directional LSTM so that
it can be used online~\cite{kim2017improved}, transferring a deep BLSTM model into a shallower one~\cite{ding2019compression},
and the posterior fusion of multiple models to improve performance~\cite{Kurata2019GuidingCP}.


\newpara\noindent{\textbf{Cross-modal distillation.}}
Our approach falls into a group of works that use networks trained on one modality to transfer knowledge to
another, in a teacher-student manner. There have been many variations on this idea, such as
using a visual recognition network (trained on RGB images) as a teacher for student networks which take
depth or optical flow~\cite{Gupta2016CrossMD}, or audio~\cite{aytar16soundnet} as inputs.
More specific examples include using 
the output of a pre-trained face emotion classifier to train a student network that can recognize emotions in speech~\cite{Albanie18}
or visual recognition of human pose to train  a network to
recognize pose from radio signals~\cite{Zhao18}.
The closest work to ours is Wei {\it et al.}~\cite{li2019improving} who 
apply cross-modal distillation from ASR for learning audio-visual speech recognition. 
An interesting finding is that the student surpasses the teacher's performance, by exploiting the extra
information available in the video modality.
However, their  method is focused on improving ASR by incorporating visual information, rather than learning
to lip read from the video signal alone, and they train the teacher model with ground truth supervision 
on the same dataset as the student one. Consequently,
their method does not apply naturally to unlabelled audio-visual data. 

 \psec\psec\psec
\section{Datasets} \label{sec:datasets} \psec
\psec
A summary of audio-visual speech datasets found in the literature is given in Table~\ref{tab:datasets}.
LRS2 and LRS3 are public audio-visual datasets that contain transcriptions but are relatively small.
LRS2 is from BBC programs and LRS3 from TED talks, and there is a domain gap between them.
Librispeech is large, transcribed, and diverse regarding the number of speakers,
but audio-only.
On the other hand VoxCeleb2, which is similar in scale, is audio-visual but lacks transcriptions.
YT31k, LSVSR and MV-LRS contain aligned
ground truth transcripts and have been used to
train state-of-the-art lip reading models \cite{makino2019recurrent, Shillingford18, Afouras19}. 
However, these datasets are not publicly available which hinders reproduction and comparison.
In this paper we focus on using only publicly available datasets. 
We use our distillation method to pretrain on VoxCeleb2 and then fine-tune and evaluate the
resulting model on LRS2 and LRS3.  

\begin{table}[t] 
\caption{ 
Statistics of modern audio-visual datasets.
{\bf Tran.:} Indicates if the dataset is labelled, i.e.\ includes aligned transcriptions; 
{\bf Mod.:} Modalities included (A=audio-only, AV=audio + video).
{\it VoxCeleb2 (clean)} refers to the subset of VoxCeleb2 we obtain after filtering according to Section~\ref{sec:datasets}.  
 }
\vspace{-8pt}
\setlength{\tabcolsep}{2.5pt}
\vspace{-10pt}                              
\begin{center}
\small
\begin{tabular}{ l r r r c c c } 
 \toprule
 {\bf Dataset}                        & {\bf\# Utter.} & {\bf\# Hours}  & {\bf Mod.} & {\bf Tran.} & {\bf Public} \\ 
 \midrule                                             
 YT31k \cite{makino2019recurrent}&      -    &  31k        & AV & \cmark & \xmark   \\  
 LSVSR \cite{Shillingford18}    &   2.9M    &  3.8k      & AV & \cmark & \xmark   \\  
 MV-LRS \cite{Chung16}          &   500k    &  775        & AV & \cmark & \xmark   \\  
 \midrule                                             
 Librispeech \cite{panayotov2015librispeech} &   292k    &  1k      & A  & \cmark & \cmark   \\  
 VoxCeleb2  \cite{Chung18a}     &   1.1M    &   2.3k            & AV & \xmark & \cmark   \\  
 \midrule                                             
 LRS2 (pre-train) \cite{Chung17}           &   96k    &  195         & AV & \cmark & \cmark   \\  
 LRS2 (main) \cite{Chung17}                &   47k    &  29          & AV & \cmark & \cmark   \\  
 LRS2 (test) \cite{Chung17}                &   1.2k   &  0.5         & AV & \cmark & \cmark   \\  
 \midrule                                             
 LRS3 (pre-train) \cite{Afouras18d}        &   132k   &  444         & AV & \cmark & \cmark   \\  
 LRS3 (train-val) \cite{Afouras18d}        &   32k    &  30          & AV & \cmark & \cmark   \\  
 LRS3 (test) \cite{Afouras18d}             &   1.3k   &   1          & AV & \cmark & \cmark   \\  
 \midrule                                             
 VoxCeleb2 (clean)              &   140k    &   334         & AV & \xmark & \cmark   \\  
 \bottomrule

\end{tabular}             
\end{center}
\label{tab:datasets}
\normalsize
\vspace{-28pt}
\end{table}


To enable the use of an unlabelled speech dataset for training lip reading models for English, we first 
filter out unsuitable videos.
For example, in VoxCeleb2, the language spoken is not always English, while the
audio in many samples can be noisy and therefore hard for an ASR model to comprehend.
We first run the trained teacher ASR model (details in section~\ref{sec:method}) to obtain
transcriptions on all the unlabelled videos. 
We then use a simple proxy to select good samples: for each utterance we calculate the
percentage of words with 4 characters or more in the ASR output that are
valid english words and
keep only the samples for which this is 90\% or more.

As a second refinement stage, we obtain transcriptions from a separate ASR model. We use a
model similar to wave2letter~\cite{liptchinsky17} trained on Librispeech. We then compare the generated
transcriptions with the ones from the teacher model and only keep an utterance when the overlap
in terms of Word Error Rate is below 28\%.
For VoxCeleb2, the above process discards a large part of the dataset,
resulting in approximately $140k$ clean utterances out of the 1M in total. 

\psec\psec
\section{Cross-modal distillation} \label{sec:method} \psec


As a {\em teacher}, we use the state-of-the-art Jasper 10x5 acoustic model \cite{li2019jasper} for ASR, 
a deep 1D-convolutional residual network. 
%
%
The {\em student} model for lip reading uses an architecture similar to the teacher's.  
More specifically, we adapt the Jasper acoustic model for lip reading as shown in Table~\ref{tab:arch}.
The input to this network are visual features extracted from a
spatio-temporal residual CNN~\cite{Stafylakis17}.

\setlength{\tabcolsep}{2.5pt}
\begin{table}[!t]
\vspace{-3pt}                              
  \caption{Architecture of Jasper-lip 5x3. To modify the Jasper model for lip-reading, we replace the first
    strided convolutional layer with a transposed convolution (stride=0.5).
}
\vspace{-5pt}                              
\label{tab:arch}
\vspace{-2pt}                              
\centering
\small
\scalebox{0.9}{
\begin{tabular}{c c c c c c} 
 \toprule
  \textbf {\# Blocks} & \textbf{Block} & \textbf{Kernel} & \textbf{\thead{\# Output\\Channels}} & \textbf{Dropout} & \textbf{\thead{\# Sub\\Blocks}} \\
 \midrule
 \multirow{2}{*}{1} & \multirow{2}{*}{Conv1} & \makecell[t]{%
 11\\%
 \textit{stride=0.5}} & \multirow{2}{*}{256} & \multirow{2}{*}{0.2} & \multirow{2}{*}{1}\\
 1 & B1 & 11 & 256 & 0.2 & 3 \\
 1 & B2 & 13 & 384 & 0.2 & 3 \\
 1 & B3 & 17 & 512 & 0.2 & 3 \\
 1 & B4 & 21 & 640 & 0.3 & 3 \\ 
 1 & B5 & 25 & 768 & 0.3 & 3 \\
 \multirow{2}{*}{1} & \multirow{2}{*}{Conv2} & \makecell[t]{%
 29\\%
 \textit{dilation=2}} & \multirow{2}{*}{896} & \multirow{2}{*}{0.4} & \multirow{2}{*}{1} \\
 1 & Conv3 & 1 & 1024  & 0.4 & 1 \\
 1 & Conv4 & 1 & \# graphemes + 1 & 0 & 1 \\
 \bottomrule
\end{tabular}
}
\normalsize
\vspace{-2pt}                              
\psec
\psec
\psec
\end{table}


\psec
\psec
\psec
\subsection{CTC loss on transcriptions} 
\psec

CTC
provides a loss function that enables training networks on sequence to sequence
tasks without the need for explicit alignment of training targets to input frames.
The CTC output token set $C'$ consists of an output grapheme alphabet $C$ augmented with a blank
symbol `$-$': $C' = C \bigcup \{-\}$.  The network consumes the input sequence
and outputs a probability distribution $p^{ctc}_t$ over $C'$ for each frame $t$.
A CTC path $\pi \in {C'}^{T}$  is a sequence of grapheme and blank labels with the same length $T$ as
the input.
Paths $\pi$ can be mapped to possible output sequences with a many-to-one function $B: {C'}^T
\rightarrow C^{\leq T} $ that removes the
blank labels and collapses repeated non-blank labels.
The probability of an output sequence $y$ given input sequence $x$ is
obtained by marginalizing over all the paths that are mapped to $y$ through B: 
$
  p(y|x) = \sum _{\pi \in B^{-1}(y)} \prod _{t=1}^{T} p^{ctc}_t(\pi(t)|x) 
$.
\cite{Graves06} computes and differentiates this sum w.r.t. the posteriors $p^{ctc}_t$ efficiently, enabling one to train the network
by minimizing the CTC loss over input-output sequence pairs $x,y^*$: 
\begin{equation}
\psec
  \mathcal{L}_{CTC}(x,y^*) = -log ( p(y^*|x) ) \nonumber
\vspace{-1pt}                              
\end{equation}

\psec
\psec
\psec
\subsection{Distillation loss} 
\psec

To distill the acoustic model into the target lip-reading model,
we minimize the KL-divergence between the teacher and student CTC posterior distributions
or, equivalently, the frame level cross-entropy loss:
\psec
\begin{equation}
  \mathcal{L}_{KD}(x_{a},x_{v}) = -\sum_{t \in T} \sum_{c \in C'} log p_{t}^{a}(c|x_a) p_{t}^{v}(c|x_v) \nonumber
\vspace{-2pt}                              
\end{equation}
where $p_{t}^{a}$ and $p_{t}^{v}$ denote the CTC posteriors for frame $t$ obtained from the teacher and student model
respectively.
This type of distillation has been used by other authors when distilling acoustic CTC models within
the same modality (audio) and is referred to as frame-wise
KD~\cite{takashima2019investigation,sak2015acoustic,kim2017improved}. 

\psec
\psec
\psec
\subsection{Combined loss} 
\psec
As shown on Fig.~\ref{fig:teaser}, given the transcription of an utterance and corresponding teacher
posteriors, we combine the CTC and KD loss terms into a common objective:
\psec
\vspace{-2pt}                              
\begin{equation}
  \mathcal{L}(x_a,x_v,y^*) = \lambda_{CTC} \mathcal{L}_{CTC}(x_v,y^*) + \lambda_{KD} \nonumber
  \mathcal{L}_{KD}(x_a,x_v)
\psec
\vspace{-2pt}                              
\end{equation}
where $\lambda_{CTC}$ and $\lambda_{KD}$ are balancing hyperparameters.

\psec\psec
\section{Experimental Setup} \label{sec:exp_setup} 
\psec

We train on the VoxCeleb2, LRS2 and LRS3 datasets and evaluate on LRS2 and LRS3 test sets (Table \ref{tab:datasets}).
In this context, we investigate the following training scenarios:

\newpara\noindent{\textbf{Full supervision.}}
We use annotated datasets only (LRS2, LRS3), and train with CTC loss on
the ground truth transcriptions, similarly to \cite{Assael16,Afouras18b}.
This is the baseline method. 

\newpara\noindent{\textbf{No supervision.}}
We do not use any ground truth transcriptions and rely solely on the transcriptions and posteriors of the
ASR teacher model for the training signal.

\newpara\noindent{\textbf{Unsupervised pre-training and fine-tuning.}}
We first pre-train the model using distillation on unlabeled data.
We then fine-tune the model on the transcribed target dataset (either LRS2 or LRS3) with full supervision.
We perform two sets of experiments in this setting:
(i) we use the ground truth annotations of all the samples in the dataset that we are fine-tuning on, or 
(ii) we only use the ground truth of the ``main'' and ``train-val'' subsets of LRS2 and LRS3
respectively (see Table \ref{tab:datasets}), which contain a small fraction of the total hours.



\psec
\psec
\psec
\subsection{Implementation details}
\psec
Our implementation is based on the Nvidia Seq2Seq framework \cite{openseq2seq}.
As a teacher, we use the 10x5 Jasper model trained on Librispeech.
To extract visual features from videos we use the publicly available visual
frontend from \cite{Afouras18b}, pre-trained on word-level lip reading.
We train the student model with the NovoGrad optimizer and
the settings of \cite{li2019jasper} on 4 GPUs with 11GB memory and a batch size of 64 on each.
We set $\lambda_{CTC}$~=~$0.1$ and $\lambda_{KD}$~=~$10$; these values were empirically determined to give similar gradient norms for each term during training.
Decoding is performed with a 8192-width beam search that uses a 6-gram language model trained on the
Librispeech corpus text.

\psec \psec 
\psec
\section{Experiments}
\psec

\begin{table}[t] 
\caption{ 
Word Error Rate \% (WER, lower is better) evaluation.
{\bf CTC:} Model trained with CTC loss.
{\bf CTC + KD:} Combined loss.
$GT$ denotes using all the ground truth transcriptions of the dataset, $ASR$
the transcriptions obtained from the teacher ASR model, and $ASR/GT$ first pre-training with the $ASR$
transcriptions and then fine-tuning with a small fraction of the ground truth ones.
{\bf Vox.:} VoxCeleb2 (clean).
$^\dagger$Trained on large non-public labelled datasets: YT31k~\cite{makino2019recurrent}, LSVSR~\cite{Shillingford18}, and
MV-LRS~\cite{Afouras19} (see Table~\ref{tab:datasets}).
$^\ddagger$Concurrent work.
 }
\vspace{-17pt}
\setlength{\tabcolsep}{2.0pt}
\begin{center}
\begin{tabular}{ l c c c | r r } 
 \toprule
 & \multicolumn{3}{c}{\textbf{Trained on}}  &  \multicolumn{2}{c}{\textbf{Evaluated on}} \\  
\addlinespace[2pt]
\textbf{Method} & Vox.  & LRS2  & LRS3     & LRS2 & LRS3  \\ 
 \midrule
 Hyb. CTC/Att. \cite{petridis2018audio}             &\xmark & GT    & \xmark  &  63.5 &   -    \\  
 TM-seq2seq$^\dagger$ \cite{Afouras19}              &\xmark & GT    &  GT     &  48.3 &  58.9  \\  
 CTC-V2P$^\dagger$ \cite{Shillingford18}              &\xmark &\xmark & \xmark  &  -    &  55.1  \\  
 RNN-T$^\dagger$ \cite{makino2019recurrent}         &\xmark &\xmark & \xmark  &  -    &  33.6  \\  
 \midrule
 Conv-seq2seq$^\ddagger$ \cite{zhang2019spatio}     &\xmark & GT    &  GT     &  \textbf{51.7} & \textbf{60.1}  \\  
 \midrule
 CTC                                      &\xmark & GT       & \xmark &  58.5  &   -    \\  
 CTC + KD                                 &\xmark & ASR      & \xmark &  58.2  &   -    \\  
 CTC + KD                                 &\xmark & ASR/GT   & \xmark &  57.9  &   -    \\  
 \midrule
 CTC                                      &\xmark & \xmark & GT       &   -   &  68.8    \\  
 CTC + KD                                 &\xmark & \xmark & ASR      &   -   &  65.6   \\  
 CTC + KD                                 &\xmark & \xmark & ASR/GT   &   -   &  65.1    \\  
 \midrule

 CTC + KD                                 & ASR & ASR    & ASR        &  54.2         & - \\  
 CTC + KD                                 & ASR & ASR/GT & ASR        &  52.2         & - \\  
 CTC + KD                               & ASR & GT       & ASR        & \textbf{51.3} & - \\  
 \midrule
 CTC + KD                                 & ASR & \xmark  & ASR        & -  & 61.7  \\  
 CTC + KD                                 & ASR & \xmark  & ASR/GT     & - &  61.0  \\ 
 CTC + KD                                 & ASR & \xmark  & GT         & - & \textbf{59.8} \\  
 \bottomrule
\end{tabular}             
\normalsize
\end{center}
\label{tab:results}
\vspace{-25pt}
\end{table}

\begin{figure}[t]
\centering 
\vspace{-10pt}
\includegraphics[width=1\linewidth]{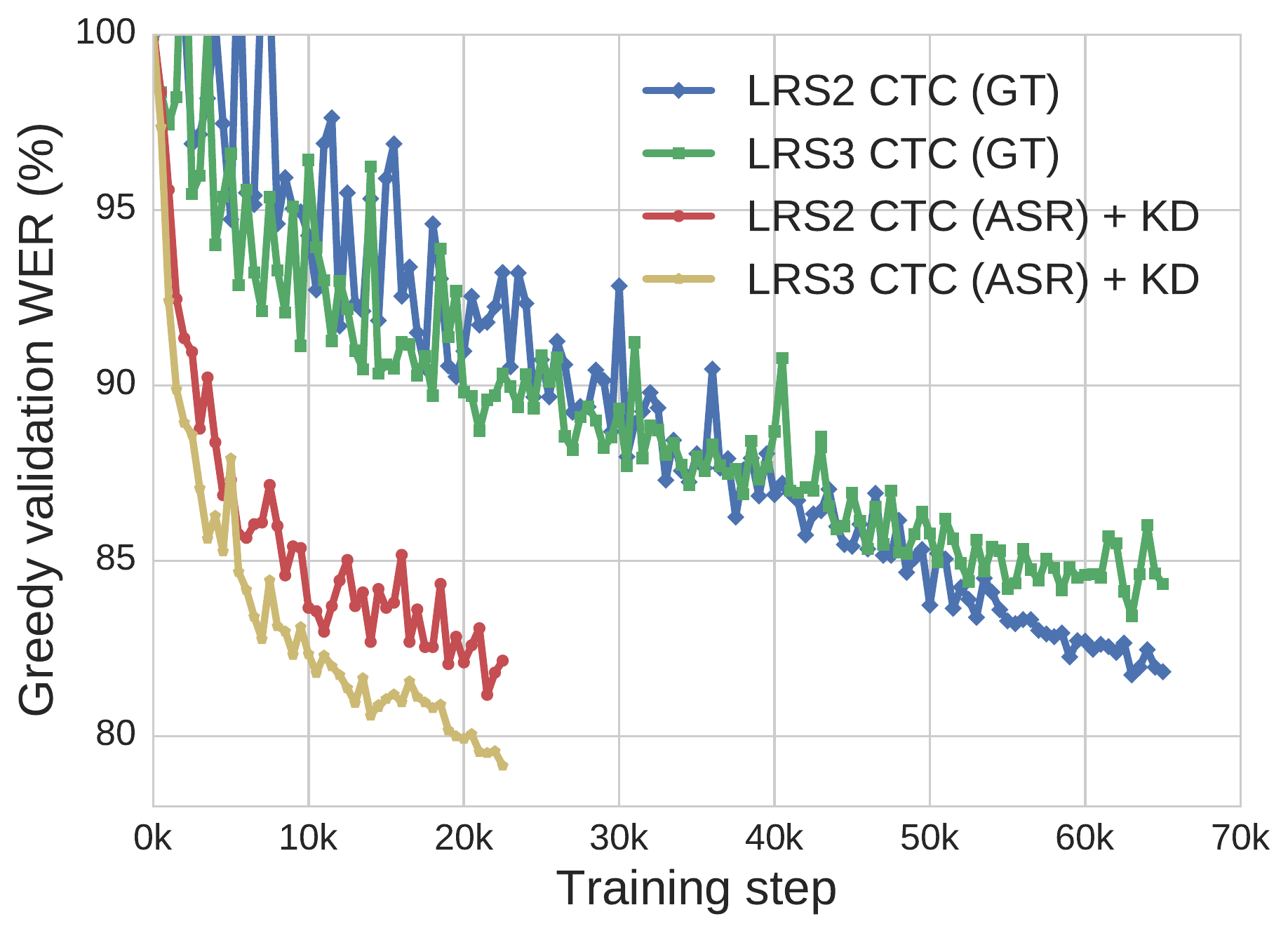}
\vspace{-15pt}
\caption{Progression of the greedy WER (validation) during
  training. Our method accelerates training significantly
compared to training with CTC alone. }
\label{fig:speed} 
\vspace{-10pt}
\psec
\psec
\end{figure}

We summarize our results in Table~\ref{tab:results}.  
The baseline method (CTC, GT) obtains $58.5\%$ WER on LRS2 and $68.8\%$ on LRS3 when trained and
evaluated on each dataset separately.  
In the same setting, and without any ground truth transcriptions, our
method achieves similar performance on LRS2 ($58.2\%$) and even better on LRS3 ($65.6\%$).
This result demonstrates that human-annotated videos are not necessary in order to effectively train lip reading models.
Fine-tuning with limited ground truth transcriptions, as described in
Section~\ref{sec:exp_setup}, reduces this to $57.9\%$ for LRS2 and $65.1\%$ for LRS3.
For training on LRS2 alone, these results outperform the previous state-of-the art
which was $63.5\%$ by~\cite{petridis2018audio}.

Using our method to train on the extra available data,
but without any ground truth transcriptions, 
we further reduce the WER to $54.2\%$ and $61.7\%$ for LRS2 and LRS3 respectively.
If we moreover fine-tune with a small amount of ground truth transcriptions, the WER drops to $52.2\%$
(LRS2) and $61.0\%$ (LRS3).
Finally, training on each dataset with full supervision after unsupervised pre-training, yields the
best results, $51.3\%$ for LRS2 and $59.8\%$ for LRS3.
Comparing these numbers to the results we obtained when training on each dataset individually, one
concludes that using extra unlabelled audio-visual speech data is indeed an effective way to boost
performance.

Distillation significantly accelerates training, even compared to using ground truth
transcriptions.
In Fig.~\ref{fig:speed} we indicatively compare the learning curves of the baseline model,
trained with CTC loss on ground truth transcriptions, and our proposed method, trained on transcriptions and posteriors from the teacher model.
Our intuition is that the acceleration is due to the distillation providing explicit alignment
information, contrary to CTC which only provides an implicit signal.


\psec
\psec
\psec
\section{Discussion and future work} \psec
\psec
In this paper we demonstrated an effective strategy to train strong models for
visual speech recognition by {\em distilling} knowledge from a pre-trained ASR model. 
This training method does not require manually annotated data
and is therefore suitable for pre-training on unlabeled datasets. 
It can be optionally fine-tuned on a small amount of annotations and achieves
performance that exceeds all existing lip reading systems aside from those trained
using proprietary data.

In concurrent work, \cite{zhang2019spatio} also obtain state-of-the-art results on LRS2 and LRS3 that are very close to ours.
We note that their improvements come from changes in the architecture, which should be orthogonal to our methodology; the two could be combined in future work for even better results.

There are many languages for which annotated data for visual speech recognition is very limited.
Since our method is applicable to any video with a talking head,
given access to a pretrained ASR model and unlabelled data for a new language,
we could naturally extend to lip reading that language.

Several authors~\cite{takashima2019investigation, Kurata2019GuidingCP, ding2019compression, sak2015acoustic} have reported difficulties
distilling acoustic models trained with CTC, stemming from the misalignment between the teacher and student spike timings. 
From the solutions proposed in the literature we only experimented with sequence-level KD~\cite{takashima2019investigation}
but did not observe any improvements.
Investigating the extent of this problem in the cross-modal distillation domain is left to future work.

The method we have proposed can be scaled to arbitrarily large amounts of data.
Given resource constraints we only utilized VoxCeleb2 and trained a relatively small network.
In future work we plan to scale up in terms of both dataset and model size to 
develop models that can match and surpass the ones trained on very large-scale annotated datasets.


\psec
\psec
\section{Acknowledgements}
\psec
Funding for this research is provided by the UK EPSRC
CDT in Autonomous Intelligent Machines and Systems, 
the Oxford-Google DeepMind Graduate Scholarship, and the EPSRC 
Programme Grant Seebibyte EP/M013774/1.



\clearpage
\bibliographystyle{IEEEbib}
{\footnotesize
\bibliography{shortstrings,vgg_local,vgg_other,mybib}

\begin{thebibliography}{10}

\bibitem{Assael16}
Yannis~M. Assael, Brendan Shillingford, Shimon Whiteson, and Nando de~Freitas,
\newblock ``Lipnet: Sentence-level lipreading,''
\newblock {\em arXiv:1611.01599}, 2016.

\bibitem{Chung16}
Joon~Son Chung and Andrew Zisserman,
\newblock ``Lip reading in the wild,''
\newblock in {\em Proc. ACCV}, 2016.

\bibitem{Chung17}
Joon~Son Chung, Andrew Senior, Oriol Vinyals, and Andrew Zisserman,
\newblock ``Lip reading sentences in the wild,''
\newblock in {\em Proc. CVPR}, 2017.

\bibitem{Shillingford18}
Brendan Shillingford, Yannis Assael, Matthew~W. Hoffman, Thomas Paine, Cían
  Hughes, Utsav Prabhu, Hank Liao, Hasim Sak, Kanishka Rao, Lorrayne Bennett,
  Marie Mulville, Ben Coppin, Ben Laurie, Andrew Senior, and Nando de~Freitas,
\newblock ``{Large-Scale Visual Speech Recognition},''
\newblock in {\em INTERSPEECH}, 2019.

\bibitem{makino2019recurrent}
Takaki Makino, Hank Liao, Yannis Assael, Brendan Shillingford, Basilio Garcia,
  Otavio Braga, and Olivier Siohan,
\newblock ``Recurrent neural network transducer for audio-visual speech
  recognition,''
\newblock in {\em IEEE Workshop on Automatic Speech Recognition and
  Understanding}, 2019.

\bibitem{Afouras18d}
Triantafyllos Afouras, Joon~Son Chung, and Andrew Zisserman,
\newblock ``{LRS3-TED}: a large-scale dataset for visual speech recognition,''
\newblock in {\em arXiv preprint arXiv:1809.00496}, 2018.

\bibitem{panayotov2015librispeech}
Vassil Panayotov, Guoguo Chen, Daniel Povey, and Sanjeev Khudanpur,
\newblock ``{Librispeech: an ASR corpus based on public domain audio books},''
\newblock in {\em Proc. ICASSP}. IEEE, 2015, pp. 5206--5210.

\bibitem{Baumann2018}
Timo Baumann, Arne K{\"o}hn, and Felix Hennig,
\newblock ``{The Spoken Wikipedia Corpus collection: Harvesting, alignment and
  an application to hyperlistening},''
\newblock {\em Language Resources and Evaluation}, 2018.

\bibitem{Chung18a}
Joon~Son Chung, Arsha Nagrani, and Andrew Zisserman,
\newblock ``{VoxCeleb2}: Deep speaker recognition,''
\newblock in {\em INTERSPEECH}, 2018.

\bibitem{ephrat2018looking}
Ariel Ephrat, Inbar Mosseri, Oran Lang, Tali Dekel, Kevin Wilson, Avinatan
  Hassidim, William~T Freeman, and Michael Rubinstein,
\newblock ``Looking to listen at the cocktail party: A speaker-independent
  audio-visual model for speech separation,''
\newblock {\em ACM Transactions on Graphics}, vol. 37, no. 4, pp. 1--11, 2018.

\bibitem{Graves06}
Alex Graves, Santiago Fern{\'a}ndez, Faustino Gomez, and J{\"u}rgen
  Schmidhuber,
\newblock ``Connectionist temporal classification: labelling unsegmented
  sequence data with recurrent neural networks,''
\newblock in {\em Proc. ICML}. ACM, 2006, pp. 369--376.

\bibitem{Stafylakis17}
Themos Stafylakis and Georgios Tzimiropoulos,
\newblock ``{Combining Residual Networks with LSTMs for Lipreading},''
\newblock in {\em Interspeech}, 2017.

\bibitem{Afouras18b}
Triantafyllos Afouras, Joon~Son Chung, and Andrew Zisserman,
\newblock ``Deep lip reading: a comparison of models and an online
  application,''
\newblock in {\em INTERSPEECH}, 2018.

\bibitem{zhang2019spatio}
Xingxuan Zhang, Feng Cheng, and Shilin Wang,
\newblock ``Spatio-temporal fusion based convolutional sequence learning for
  lip reading,''
\newblock in {\em Proc. ICCV}, 2019, pp. 713--722.

\bibitem{petridis2018audio}
Stavros Petridis, Themos Stafylakis, Pingchuan Ma, Georgios Tzimiropoulos, and
  Maja Pantic,
\newblock ``{Audio-Visual Speech Recognition with a Hybrid CTC/Attention
  Architecture},''
\newblock in {\em IEEE Spoken Language Technology Workshop}, 2018, pp.
  513--520.

\bibitem{Afouras19}
Triantafyllos Afouras, Joon~Son Chung, Andrew Senior, Oriol Vinyals, and Andrew
  Zisserman,
\newblock ``Deep audio-visual speech recognition,''
\newblock {\em IEEE PAMI}, 2019.

\bibitem{Kannan17}
Anjuli Kannan, Yonghui Wu, Patrick Nguyen, Tara~N. Sainath, Zhifeng Chen, and
  Rohit Prabhavalkar,
\newblock ``{An analysis of incorporating an external language model into a
  sequence-to-sequence model},''
\newblock in {\em Proc. ICASSP}, 2018.

\bibitem{Maas15}
Andrew~L. Maas, Ziang Xie, Dan Jurafsky, and Andrew~Y. Ng,
\newblock ``Lexicon-free conversational speech recognition with neural
  networks,''
\newblock in {\em Proceedings the North American Chapter of the Association for
  Computational Linguistics}, 2015.

\bibitem{hinton2015distilling}
Geoffrey Hinton, Oriol Vinyals, and Jeff Dean,
\newblock ``Distilling the knowledge in a neural network,''
\newblock in {\em NIPS Deep Learning and Representation Learning Workshop},
  2015.

\bibitem{ba2014deep}
Jimmy Ba and Rich Caruana,
\newblock ``Do deep nets really need to be deep?,''
\newblock in {\em NeurIPS}, 2014, pp. 2654--2662.

\bibitem{li2014learning}
Jinyu Li, Rui Zhao, Jui-Ting Huang, and Yifan Gong,
\newblock ``{Learning small-size DNN with output-distribution-based
  criteria},''
\newblock in {\em INTERSPEECH}, 2014.

\bibitem{kim2016sequence}
Yoon Kim and Alexander~M Rush,
\newblock ``Sequence-level knowledge distillation,''
\newblock in {\em Proc. EMNLP}, 2016.

\bibitem{kim2019knowledge}
Ho-Gyeong Kim, Hwidong Na, Hoshik Lee, Jihyun Lee, Tae~Gyoon Kang, Min-Joong
  Lee, and Young~Sang Choi,
\newblock ``Knowledge distillation using output errors for self-attention
  end-to-end models,''
\newblock in {\em Proc. ICASSP}. IEEE, 2019, pp. 6181--6185.

\bibitem{kim2017improved}
Suyoun Kim, Michael~L. Seltzer, Jinyu Li, and Rui Zhao,
\newblock ``Improved training for online end-to-end speech recognition
  systems,'' 2017.

\bibitem{ding2019compression}
Haisong Ding, Kai Chen, and Qiang Huo,
\newblock ``{Compression of CTC-Trained Acoustic Models by Dynamic Frame-Wise
  Distillation or Segment-Wise N-Best Hypotheses Imitation},''
\newblock in {\em INTERSPEECH}, 2019, pp. 3218--3222.

\bibitem{Kurata2019GuidingCP}
Gakuto Kurata and Kartik Audhkhasi,
\newblock ``{Guiding CTC Posterior Spike Timings for Improved Posterior Fusion
  and Knowledge Distillation},''
\newblock in {\em INTERSPEECH}, 2019.

\bibitem{Gupta2016CrossMD}
Saurabh Gupta, Judy Hoffman, and Jitendra Malik,
\newblock ``Cross modal distillation for supervision transfer,''
\newblock in {\em Proc. CVPR}, 2016.

\bibitem{aytar16soundnet}
Yusuf Aytar, Carl Vondrick, and Antonio Torralba,
\newblock ``Soundnet: Learning sound representations from unlabeled video,''
\newblock in {\em NeurIPS}, 2016.

\bibitem{Albanie18}
Samuel Albanie, Arsha Nagrani, Andrea Vedaldi, and Andrew Zisserman,
\newblock ``Emotion recognition in speech using cross-modal transfer in the
  wild,''
\newblock in {\em Proc. ACMM}, 2018.

\bibitem{Zhao18}
Mingmin Zhao, Tianhong Li, Mohammad Abu~Alsheikh, Yonglong Tian, Hang Zhao,
  Antonio Torralba, and Dina Katabi,
\newblock ``Through-wall human pose estimation using radio signals,''
\newblock in {\em Proc. CVPR}, June 2018.

\bibitem{li2019improving}
Wei Li, Sicheng Wang, Ming Lei, Sabato~Marco Siniscalchi, and Chin-Hui Lee,
\newblock ``Improving audio-visual speech recognition performance with
  cross-modal student-teacher training,''
\newblock in {\em Proc. ICASSP}. IEEE, 2019, pp. 6560--6564.

\bibitem{liptchinsky17}
Vitaliy Liptchinsky, Gabriel Synnaeve, and Ronan Collobert,
\newblock ``Letter-based speech recognition with gated {ConvNets},''
\newblock {\em CoRR}, vol. abs/1712.09444, 2017.

\bibitem{li2019jasper}
Jason Li, Vitaly Lavrukhin, Boris Ginsburg, Ryan Leary, Oleksii Kuchaiev,
  Jonathan~M. Cohen, Huyen Nguyen, and Ravi~Teja Gadde,
\newblock ``Jasper: An end-to-end convolutional neural acoustic model,''
\newblock in {\em INTERSPEECH}, 2019.

\bibitem{takashima2019investigation}
Ryoichi Takashima, Li~Sheng, and Hisashi Kawai,
\newblock ``{Investigation of Sequence-level Knowledge Distillation Methods for
  CTC Acoustic Models},''
\newblock in {\em Proc. ICASSP}. IEEE, 2019, pp. 6156--6160.

\bibitem{sak2015acoustic}
Andrew Senior, Ha{\c{s}}im Sak, F{\'e}lix de~Chaumont~Quitry, Tara Sainath,
  Kanishka Rao, et~al.,
\newblock ``{Acoustic modelling with CD-CTC-sMBR LSTM RNNs},''
\newblock in {\em IEEE Workshop on Automatic Speech Recognition and
  Understanding}. IEEE, 2015, pp. 604--609.

\bibitem{openseq2seq}
Oleksii Kuchaiev, Boris Ginsburg, Igor Gitman, Vitaly Lavrukhin, Jason Li,
  Huyen Nguyen, Carl Case, and Paulius Micikevicius,
\newblock ``{Mixed-Precision Training for NLP and Speech Recognition with
  OpenSeq2Seq},'' 2018.

\end{thebibliography}
}

\end{document}